\def\year{2019}\relax
\documentclass[letterpaper]{article} 
\usepackage{aaai19}  
\usepackage{times}  
\usepackage{helvet}  
\usepackage{courier}  
\usepackage{url}  
\usepackage{graphicx}  

\usepackage{latexsym}
\usepackage{amsmath}
\usepackage{algorithm, algorithmicx, algpseudocode}
\usepackage{multirow,array}
\usepackage{enumitem}
\usepackage{makecell}
\usepackage{pinyin}
\usepackage[normalem]{ulem}

\newcommand{\citet}[1]{\citeauthor{#1}~\shortcite{#1}}

\newcommand{\tabincell}[2]{\begin{tabular}{@{}#1@{}}#2\end{tabular}} 

\begin{document}
%

\title{Regularizing Neural Machine Translation by Target-bidirectional Agreement}

\author{}

\author{ Zhirui Zhang$^\ddag$\thanks{Contribution during internship at Microsoft.}, Shuangzhi Wu$^\natural$, Shujie Liu$^\S$, Mu Li$^\S$, Ming Zhou$^\S$, Tong  Xu$^\ddag$\thanks{Corresponding Author.}\\
  $^\ddag$University of Science and Technology of China, Hefei, China\\
  $^\natural$Harbin Institute of Technology, Harbin, China \ \ $^\S$Microsoft Research Asia \\
    $^\ddag$zrustc11@gmail.com \  $^\ddag$tongxu@ustc.edu.cn \\ 
    $^{\natural\S}$\{v-shuawu,shujliu,muli,mingzhou\}@microsoft.com \\
}

\maketitle
\begin{abstract}
		
Although Neural Machine Translation (NMT) has achieved remarkable progress in the past several years, most NMT systems still suffer from a fundamental shortcoming as in other sequence generation tasks: errors made early in generation process are fed as inputs to the model and can be quickly amplified, harming subsequent sequence generation.
To address this issue, we propose a novel model regularization method for NMT training, which aims to improve the agreement between translations generated by left-to-right (L2R) and right-to-left (R2L) NMT decoders.
This goal is achieved by introducing two Kullback-Leibler divergence regularization terms into the NMT training objective to reduce the mismatch between output probabilities of L2R and R2L models.
In addition, we also employ a joint training strategy to allow L2R and R2L models to improve each other in an interactive update process.
Experimental results show that our proposed method significantly outperforms state-of-the-art baselines on Chinese-English and English-German translation tasks.
		
\end{abstract}

\section{Introduction}

Neural Machine Translation (NMT) \cite{cho2014learning,sutskever2014sequence,Bahdanau2014NeuralMT} has seen the rapid development in the past several years, from catching up with Statistical Machine Translation (SMT) \cite{Koehn2003StatisticalPT,chiang2007hierarchical} to outperforming it by significant margins on many languages \cite{Sennrich2016NeuralMT,Wu2016GooglesNM,Tu2016ModelingCF,Eriguchi2016TreetoSequenceAN,Wang2017DeepNM,Vaswani2017AttentionIA}.
In a conventional NMT model, an encoder first transforms the source sequence into a sequence of intermediate hidden vector representations, based on which, a decoder generates the target sequence word by word. 

Due to the autoregressive structure, current NMT systems usually suffer from the so-called exposure bias problem \cite{bengio2015scheduled}: 
during inference, true previous target tokens are unavailable and replaced by tokens generated by the model itself, thus mistakes made early can mislead subsequent translation, yielding unsatisfied translations with good prefixes but bad suffixes (shown in Table \ref{table:introduction}).
Such an issue can become severe as sequence length increases.


\begin{table}[t] 
\centering
\begin{tabular}{r|l}
\hline
Input & \tabincell{l}{\zhi1\chi2\zhe3\ \men1\ \biao3\shi4,\ \zhe4\ \liang3\tiao2\ \sui4\dao4\ \jiang1\\\shi3\ \huan2\jing4\ \shou4\yi4\ \bing4\ \bang1\zhu4\ \jia1\li4\fu2\ni2\ya4\zhou1\\\que4\bao3\ \gong4\shui3\ \geng4\jia1\ \an1\quan2.}       \\ \hline
Ref. & \tabincell{l}{Supporters say the two tunnels will benefit the \\ environment and help California ensure the water \\ supply is safer. }     \\ \hline
L2R & \tabincell{l}{Supporters say these two tunnels will benefit the \\ environment and \uwave{to help a secure water supply} \\ \uwave{in California.}  }         \\ \hline
R2L & \tabincell{l}{\uwave{Supporter say the tunnel} will benefit the environ-\\ment and help California ensure the water supply\\ is more secure. }    \\ \hline
\end{tabular}
\caption{Example of an unsatisfied translation generate by a left-to-right (L2R) decoder and a right-to-left (R2L) decoder.}
\label{table:introduction}
\end{table}

To address this problem, one line of research attempts to reduce the inconsistency between training and inference so as to improve the robustness when giving incorrect previous predictions, such as designing sequence-level objectives or adopting reinforcement learning approaches \cite{ranzato2015sequence,shen-EtAl:2016:P16-1,wiseman-rush:2016:EMNLP2016}. Another line tries to leverage  a complementary NMT model that generates target words from right to left (R2L) to distinguish unsatisfied translation results from a n-best list generated by the L2R model \cite{liu2016agreement,Wang2017SogouNM}.

In their work, the R2L NMT model is only used to re-rank the translation candidates generated by L2R model, while the candidates in the n-best list still suffer from the exposure bias problem and limit the room for improvement. Another problem is that, the complementary R2L model tends to generate  translation results with good suffixes and bad prefixes, due to the same exposure bias problem, as shown in Table \ref{table:introduction}.  Similar with using the R2L model to augment the L2R model, the L2R model can also be leveraged to improve the R2L model.


Instead of re-ranking the n-best list, we try to take consideration of the agreement between the L2R and R2L models into both of their training objectives, hoping that the agreement information can help to learn better models integrating their advantages to generate translations with good prefixes and good suffixes.
To this end, we introduce two Kullback-Leibler (KL) divergences between the probability distributions defined by L2R and R2L models into the NMT training objective as regularization terms. 
Thus, we can not only maximize the likelihood of training data but also minimize the L2R and R2L model divergence at the same time, in which the latter one severs as a measure of exposure bias problem of the currently evaluated model. 
With this method, the L2R model can be enhanced using the R2L model as a helper system, and the R2L model can also be improved with the help of L2R model. We integrate the optimization of R2L and L2R models into a joint training framework, in which they act as helper systems for each other, and both models achieve further improvements with an interactive update process.

Our experiments are conducted on Chinese-English and English-German translation tasks, and demonstrate that our proposed method significantly outperforms state-of-the-art baselines.

\section{Neural Machine Translation}

Neural Machine Translation (NMT) is an end-to-end framework to directly model the conditional probability \(P(y|x)\) of target translation \(y=(y_1, y_2, ...\, , y_{T'})\) given source sentence \( x=(x_1, x_2, ...\, , x_T ) \).
In practice, NMT systems are usually implemented with an attention-based encoder-decoder architecture.
The encoder reads the source sentence \(x\) and transforms it into a sequence of intermediate hidden vectors \( h = (h_1, h_2,...\, ,h_T)\) using a neural network.
Given the hidden state \(h\), the decoder generates target translation \(y\) with another neural network that jointly learns language and alignment models.

The structure of neural networks first employs recurrent neural networks (RNN) or its variants - Gated Recurrent Unit \cite{cho2014learning} and Long Short-Term Memory \cite{hochreiter1997long} networks. Recently, two additional architectures have been proposed, improving not only parallelization but also the state-of-the-art result: the fully convolution model \cite{Gehring2017ConvolutionalST} and the self-attentional transformer \cite{Vaswani2017AttentionIA}. 

For model training, given a parallel corpus \( D = \{ (x^{(n)}, y^{(n)}) \}_{n=1}^{N} \), 
the standard training objective in NMT is to maximize the likelihood of the training data:
\begin{equation}
	L(\theta) = \sum_{n=1}^{N} \log{ P(y^{(n)}| x^{(n)};\theta)}
	\label{equ:MLE-loss}
\end{equation}
where \( P(y|x; \theta) \) is the neural translation model and \( \theta \) is the model parameter.

One big problem of the model training is that the history of any target word is correct and has been observed in the training data,
but during inference, all the target words are predicted and may contain mistakes, which are fed as inputs to the model and quickly accumulated along with the sequence generation.
This is called the exposure bias problem \cite{bengio2015scheduled}.

\section{Our Approach}

To deal with the exposure bias problem, we try to maximize the agreement between translations from L2R and R2L NMT models, and divide the NMT training objective into two parts: the standard maximum likelihood of training data, and the regularization terms that indicate the divergence of L2R and R2L models based on the current model parameter.
In this section, we will start with basic model notations, followed by discussions of model regularization terms and efficient gradient approximation methods. In the last part, we show that the L2R and R2L NMT models can be jointly improved to achieve even better results.

\subsection{Notations}

Given source sentence \( x=(x_1, x_2, ...\, , x_T ) \) and its target translation \( y=(y_1, y_2, ...\, , y_{T'})\), let \( P(y|x; \overrightarrow{\theta} ) \) and \( P(y|x; \overleftarrow{\theta} ) \) be L2R and R2L translation models, in which \( \overrightarrow{\theta} \) and \( \overleftarrow{\theta} \) are corresponding model parameters.
Specifically, L2R translation model can be decomposed as \( P(y|x; \overrightarrow{\theta} ) = \prod_{t=1}^{T'} P(y_t | y_{<t}, x ; \overrightarrow{\theta} ) \), which means L2R model adopts previous targets \( y_1,\dots, y_{t-1} \) as history to predict the current target \(y_{t} \) at each step \( t \), while the R2L translation model can similarly be decomposed as \( P(y|x; \overleftarrow{\theta} )  = \prod_{t=T'}^{1} P(y_t | y_{>t}, x ; \overleftarrow{\theta} ) \) and employs later targets \( y_{t+1},\dots, y_{T'} \) as history to predict current target \(y_{t} \) at each step \( t \).


\subsection{NMT Model Regularization}

Since L2R and R2L models are different chain decompositions of the same translation probability, output probabilities of the two models should be identical:
\begin{equation}
	\begin{aligned}
		\log P(y|x;\overrightarrow{\theta} ) & = \sum_{t=1}^{T'} \log P(y_t | y_{<t}, x ; \overrightarrow{\theta} ) \\
		& = \sum_{t=T'}^{1} \log P(y_t | y_{>t}, x ; \overleftarrow{\theta} ) \\
		& = \log P(y|x; \overleftarrow{\theta} ) 
	\end{aligned}
	\label{equ:identical-equ}
\end{equation}
However, if these two models are optimized separately by maximum likelihood estimation (MLE), there is no guarantee that the above equation will hold.
To satisfy this constraint, we introduce two Kullback-Leibler (KL) divergence regularization terms into the MLE training objective (Equation \ref{equ:MLE-loss}). For L2R model, the new training objective is:
\begin{equation}
	\begin{aligned}
		L(\overrightarrow{\theta}) & = \sum_{n=1}^{N}  \log{ P(y^{(n)}| x^{(n)};\overrightarrow{\theta} )} \\
		& - \lambda \sum_{n=1}^{N} \text{KL}( P(y|x^{(n)}; \overleftarrow{\theta} ) || P(y|x^{(n)}; \overrightarrow{\theta} ) ) \\
		& - \lambda \sum_{n=1}^{N} \text{KL}( P(y|x^{(n)}; \overrightarrow{\theta} ) || P(y|x^{(n)}; \overleftarrow{\theta} ) ) 
	\end{aligned}
	\label{equ:regularization}
\end{equation}
where \( \lambda \) is a hyper-parameter for regularization terms.
These regularization terms are 0 when Equation \ref{equ:identical-equ} holds, otherwise regularization terms will guide the training process to reduce the disagreement between L2R and R2L models.

Unfortunately, it is impossible to calculate entire gradients of this objective function, since we need to sum over all translation candidates in an exponential search space for KL divergence.
To alleviate this problem, we follow \citet{shen-EtAl:2016:P16-1} to approximate the full search space with a sampled sub-space and then design an efficient KL divergence approximation algorithm.
Specifically, we derive the gradient calculation equation based on the definition of KL divergence, and then design proper sampling methods for two different KL divergence regularization terms.

For \( \text{KL}( P(y|x^{(n)}; \overleftarrow{\theta} ) || P(y|x^{(n)}; \overrightarrow{\theta} )) \), according to the definition of KL divergence, we have  
\begin{equation}
	\begin{aligned}
		& \text{KL}( P(y|x^{(n)}; \overleftarrow{\theta} ) || P(y|x^{(n)}; \overrightarrow{\theta} ))   \\
		= & \sum_{y \in Y(x^{(n)}) }P(y|x^{(n)}; \overleftarrow{\theta} ) \log \frac{P(y|x^{(n)}; \overleftarrow{\theta} )}{P(y|x^{(n)}; \overrightarrow{\theta} )}
	\end{aligned}
\end{equation}
where \( Y(x^{(n)}) \) is a set of all possible candidate translations for the source sentence \( x^{(n)} \).
Since \( P(y|x^{(n)}; \overleftarrow{\theta} ) \) is irrelevant to parameter \( \overrightarrow{\theta} \), the partial derivative of this KL divergence with respect to \( \overrightarrow{\theta} \) can be written as
\begin{equation}
	\begin{aligned}
		& \frac{\partial \text{KL}( P(y|x^{(n)}; \overleftarrow{\theta} ) || P(y|x^{(n)}; \overrightarrow{\theta} )) }{\partial \overrightarrow{\theta} } \\
		= & -\sum_{y \in Y(x^{(n)}) } P(y|x^{(n)}; \overleftarrow{\theta} ) \frac{\partial \log P(y|x^{(n)}; \overrightarrow{\theta} ) }{\partial \overrightarrow{\theta}} \\
		= & - \text{E}_{y \sim P(y|x^{(n)}; \overleftarrow{\theta} ) } \frac{\partial \log P(y|x^{(n)}; \overrightarrow{\theta} ) }{\partial \overrightarrow{\theta}}
	\end{aligned}
	\label{equ:regularization1-gradient}
\end{equation}
in which \( \frac{\partial \log P(y|x^{(n)}; \overrightarrow{\theta} ) }{\partial \overrightarrow{\theta}} \) are the gradients specified with a standard sequence-to-sequence NMT network. The expectation $\text{E}_{y \sim P(y|x^{(n)}; \overleftarrow{\theta} ) }$ can be approximated with samples from the R2L model $P(y|x^{(n)}; \overleftarrow{\theta} )$.
Therefore, minimizing this regularization term is equal to maximizing the log-likelihood on the pseudo sentence pairs sampled from the R2L model.

For \( \text{KL}( P(y|x^{(n)}; \overrightarrow{\theta} ) || P(y|x^{(n)}; \overleftarrow{\theta} ) ) \), similarly we have  
\begin{equation}
	\begin{aligned}
		& \text{KL}( P(y|x^{(n)}; \overrightarrow{\theta} ) || P(y|x^{(n)}; \overleftarrow{\theta} ))   \\
		= & \sum_{y \in Y(x^{(n)}) }P(y|x^{(n)}; \overrightarrow{\theta} ) \log \frac{P(y|x^{(n)}; \overrightarrow{\theta} )}{P(y|x^{(n)}; \overleftarrow{\theta} )}
	\end{aligned}
\end{equation}
The partial derivative of this KL divergence with respect to \( \overrightarrow{\theta} \) is calculated as follows:
\begin{equation}
	\begin{aligned}
		& \frac{\partial \text{KL}( P(y|x^{(n)}; \overrightarrow{\theta} ) || P(y|x^{(n)}; \overleftarrow{\theta} )) }{\partial \overrightarrow{\theta} } \\
		= & - \text{E}_{y \sim P(y|x^{(n)}; \overrightarrow{\theta} ) } \\
		&  \left( \log \frac{P(y|x^{(n)}; \overleftarrow{\theta} )}{ P(y|x^{(n)}; \overrightarrow{\theta} )} 
		\frac{\partial \log P(y|x^{(n)}; \overrightarrow{\theta} ) }{\partial \overrightarrow{\theta}} \right)
	\end{aligned}
	\label{equ:regularization2-gradient}
\end{equation}
Similarly, we use sampling for the calculation of expectation $\text{E}_{y \sim P(y|x^{(n)}; \overrightarrow{\theta} ) }$. There are two differences in Equation \ref{equ:regularization2-gradient} compared with Equation \ref{equ:regularization1-gradient}: 1) Pseudo sentence pairs are not sampled from the R2L model ($P(y|x^{(n)}; \overleftarrow{\theta} )$), but from the L2R model itself ($P(y|x^{(n)}; \overrightarrow{\theta} )$); 2) \( \log \frac{P(y|x^{(n)}; \overleftarrow{\theta} )}{ P(y|x^{(n)}; \overrightarrow{\theta} )} \) is used as weight to penalize incorrect pseudo pairs.


\begin{algorithm}[t]
	\caption{Training Algorithm for L2R Model}
	\label{alg:regularization}
	\hspace*{\algorithmicindent} \textbf{Input:} Bilingual Data \( D =\{ (x^{(n)}, y^{(n)}) \}_{n=1}^{N} \)  \\
	\hspace*{\algorithmicindent} \qquad \quad R2L Model \( P(y|x; \overleftarrow{\theta} ) \)  \\
	\hspace*{\algorithmicindent} \textbf{Output:} L2R Model \( P(y|x; \overrightarrow{\theta} ) \) 
	\begin{algorithmic}[1]
		\Procedure{training process}{}
		\While{Not Converged}
		\State Sample sentence pairs \( (x^{(n)}, y^{(n)}) \) from bilingual data \( D\);
		\State Generate \(m\) translation candidates \( y_1^{1},\dots,y_m^{1} \) for \( x^{(n)} \) by  translation model \( P(y|x; \overleftarrow{\theta} ) \) and build pseudo sentence pairs \( \{ (x^{(n)}, y_{i}^{1}) \}_{i=1}^{m} \);
		\State Generate \(m\) translation candidates \( y_1^{2},\dots,y_m^{2} \) for \( x^{(n)} \) by  translation model \( P(y|x; \overrightarrow{\theta} ) \) and build pseudo sentence pairs \( \{ (x^{(n)}, y_{i}^{2}) \}_{i=1}^{m} \) weighted with \( \log \frac{P(y_i^2|x^{(n)}; \overleftarrow{\theta} )}{ P(y_i^2|x^{(n)}; \overrightarrow{\theta} )} \);
		\State  Update \( P(y|x; \overrightarrow{\theta} ) \) with Equation \ref{equ:gradient} given original data (\( (x^{(n)}, y^{(n)}) \)) and two synthetic data  (\( \{ (x^{(n)}, y_{i}^{1}) \}_{i=1}^{m} \) and \( \{ (x^{(n)}, y_{i}^{2}) \}_{i=1}^{m} \)).
		\EndWhile
		\EndProcedure
	\end{algorithmic}
\end{algorithm}

To sum up, the partial derivative of objective function \( L(\overrightarrow{\theta}) \) with respect to \( \overrightarrow{\theta} \) can be approximately written as follows:
\begin{equation}
	\begin{aligned}
		\frac{ \partial L(\overrightarrow{\theta}) }{ \partial \overrightarrow{\theta}} & = \sum_{n=1}^{N} \frac{ \partial  \log{ P(y^{(n)}| x^{(n)};\overrightarrow{\theta} )} }{ \partial \overrightarrow{\theta} }  \\
		& + \lambda \sum_{n=1}^{N} \sum_{y \sim P(y|x^{(n)}; \overleftarrow{\theta} )} \frac{\partial \log P(y|x^{(n)}; \overrightarrow{\theta} ) }{\partial \overrightarrow{\theta}} \\
		& + \lambda \sum_{n=1}^{N} \sum_{y \sim P(y|x^{(n)}; \overrightarrow{\theta} )} \\
		& \left( \log \frac{P(y|x^{(n)}; \overleftarrow{\theta} )}{ P(y|x^{(n)}; \overrightarrow{\theta} )} 
		\frac{\partial \log P(y|x^{(n)}; \overrightarrow{\theta} ) }{\partial \overrightarrow{\theta}} \right)
	\end{aligned}
	\label{equ:gradient}
\end{equation}
The overall training  is shown in Algorithm \ref{alg:regularization}.

\subsection{Joint Training for Paired NMT Models}
In practice, due to the imperfection of R2L model, the agreement between L2R and R2L models sometimes may mislead L2R model training.
On the other hand, due to the symmetry of L2R and R2L models, L2R model can also serve as the discriminator to punish bad translation candidates generated from R2L model.
Similarly, the objective function of the R2L model can be defined as follow:
\begin{equation}
	\label{equ:regularization-r2l}
	\begin{aligned}
		L(\overleftarrow{\theta}) & = \sum_{n=1}^{N}  \log{ P(y^{(n)}| x^{(n)};\overleftarrow{\theta} )} \\
		& - \lambda \sum_{n=1}^{N} \text{KL}( P(y|x^{(n)}; \overrightarrow{\theta} ) || P(y|x^{(n)}; \overleftarrow{\theta} ) )  \\
		& - \lambda \sum_{n=1}^{N} \text{KL}( P(y|x^{(n)}; \overleftarrow{\theta} ) || P(y|x^{(n)}; \overrightarrow{\theta} ) ) 
	\end{aligned}
\end{equation}
The corresponding training procedure is similar with Algorithm \ref{alg:regularization}.

\begin{figure}[t] 
	\begin{center}
	    \includegraphics[width = 3.2in]{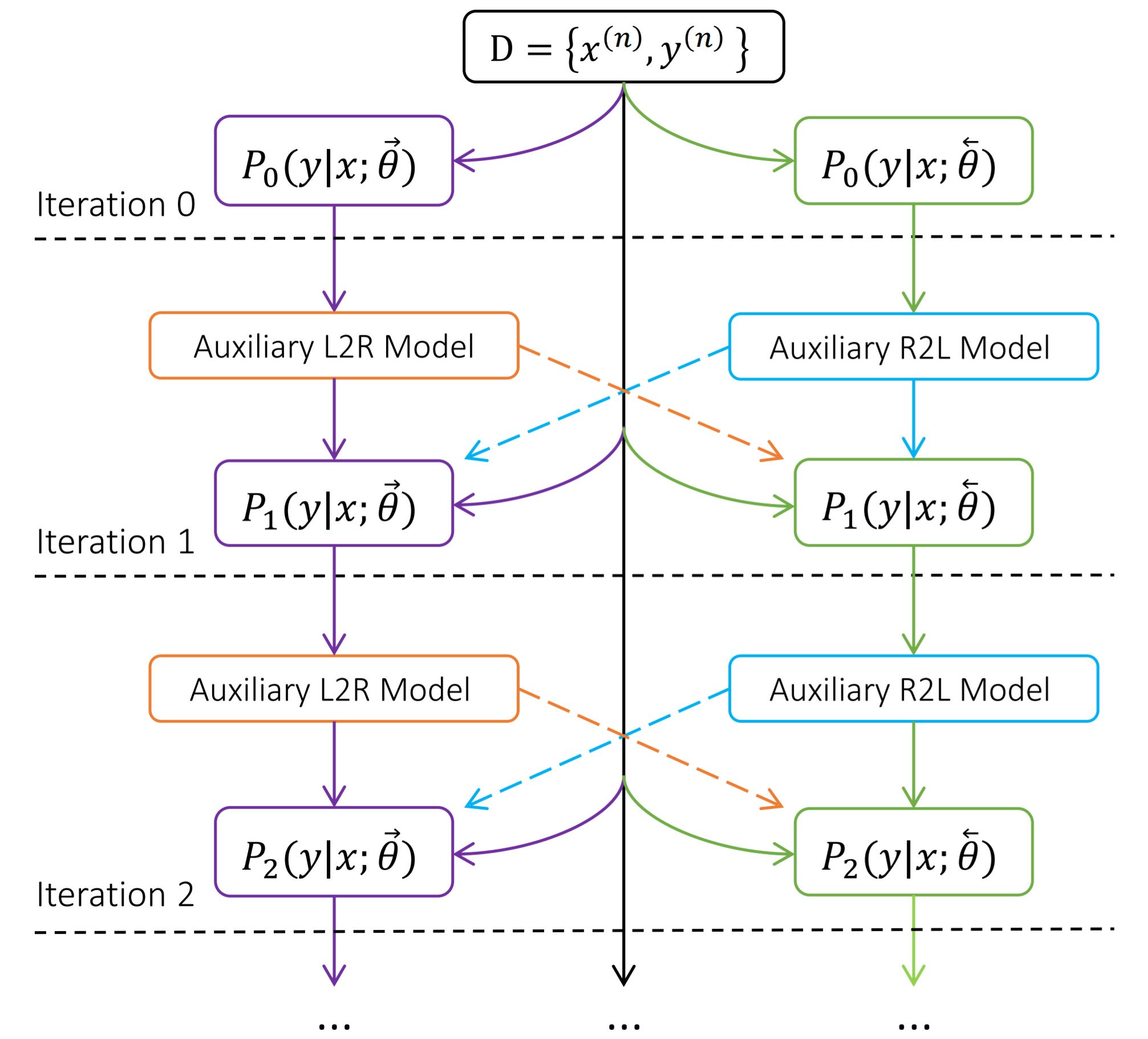}
	\end{center}
	\vspace{-8pt}
	\caption{Illustration of joint training of NMT models in two directions (L2R model $P(y| x;\overrightarrow{\theta} )$ and R2L model $P(y| x;\overleftarrow{\theta} )$).}
	\label{fig:joint-training}
\end{figure} 

Based on the above, L2R and R2L models can act as helper systems for each other in a joint training process: the L2R model \( P(y| x;\overrightarrow{\theta} ) \) is used as auxiliary system to regularize R2L model \( P(y| x;\overleftarrow{\theta} ) \), and the R2L model \( P(y| x;\overleftarrow{\theta} ) \) is used as auxiliary system to regularize L2R model \( P(y| x;\overrightarrow{\theta} ) \). 
This training process can be iteratively carried out to obtain further improvements because after each iteration both L2R and R2L models are expected to be improved with regularization method.

To simultaneously optimize these two models, we design a novel training algorithm with the overall training objective defined as the sum of objectives in both directions:
\begin{equation}
	\begin{aligned}
		L(\theta) & =  L(\overrightarrow{\theta}) + L(\overleftarrow{\theta}) 
	\end{aligned}
\end{equation}
As illustrated in Figure \ref{fig:joint-training}, the whole training process contains two major steps: pre-training and joint-training.
First, given parallel corpora \(\text{D}=\{(x^{(n)},y^{(n)})\}_{n=1}^{N} \), we pre-train both L2R and R2L models with MLE principle.
Next, based on pre-trained models, we jointly optimize L2R and R2L models with an iterative process. 
In each iteration, we fix R2L model and use it as a helper to optimize L2R model with Equation \ref{equ:regularization}, and at the same time, we fix L2R model and use it as a helper to optimize R2L model with Equation \ref{equ:regularization-r2l}.
The iterative training continues until the performance on development set does not increase.

\section{Experiments}

\subsection{Setup}

To examine the effectiveness of our proposed approach, we conduct experiments on three datasets, including NIST OpenMT for Chinese-English, WMT17 for English-German and Chinese-English. 
In all experiments, we use BLEU \cite{papineni2002bleu} as the automatic metric for translation evaluation.

\paragraph{Datasets.} 
For NIST OpenMT's Chinese-English translation task, we select our training data from LDC corpora,\footnote{The corpora include LDC2002E17, LDC2002E18, LDC2003E07, LDC2003E14, LDC2005E83, LDC2005T06, LDC2005T10, LDC2006E17, LDC2006E26, LDC2006E34, LDC2006E85, LDC2006E92, LDC2006T06, LDC2004T08, LDC2005T10} which consists of 2.6M sentence pairs with 65.1M Chinese words and 67.1M English words respectively.
Any sentence longer than 80 words is removed from training data.
The NIST OpenMT 2006 evaluation set is used as validation set, and NIST 2003, 2005, 2008, 2012 datasets as test sets.
We limit the vocabulary to contain up to 50K most frequent words on both source and target sides, and convert remaining words into the {\tt <unk>} tokens.
In decoding period, we follow \citet{LuongACL2015} to handle the {\tt <unk>} replacement.

For WMT17's English-German translation task, 
we use the pre-processed training data provided by the task organizers.\footnote{http://data.statmt.org/wmt17/translation-task/preprocessed\\/de-en/ }
The training data consists of 5.8M sentence pairs with 141M English words and 134M German words respectively.
We use the newstest2016 as the validation set and the newstest2017 as the test set.
The maximal sentence length is set as 128.
For vocabulary, we use 37K sub-word tokens based on Byte Pair Encoding (BPE) \cite{Sennrich2016NeuralMT}. 

\begin{table*}[t]
	\centering
	\begin{tabular}{l|c||c|c|c|c|c}
		\hline
		System                       & NIST2006  & NIST2003  & NIST2005  & NIST2008  & NIST2012  & Average \\ \hline \hline
		Transformer     & 44.33 & 45.69 & 43.94 & 34.80 & 32.63 & 40.28   \\ \hline
		Transformer+MRT & 45.21  & 46.60  & 45.11  & 36.77 & 34.78 & 41.69 \\ \hline
		Transformer+JS & 45.04 & 46.32 & 44.58 & 36.81 & 35.02 & 41.51   \\ \hline
		Transformer+RT              & \textbf{46.14} & \textbf{48.28} & \textbf{46.24} & \textbf{38.07} & \textbf{36.31} & \textbf{43.01}  \\ \hline
	\end{tabular}
	\caption{Case-insensitive BLEU scores (\%) for Chinese-English translation on NIST datasets. The ``Average" denotes the average BLEU score of all datasets in the same setting.}
	\label{table:chinese-english-NIST}
\end{table*}

For WMT17's Chinese-English translation task,
we use all the available parallel data, which consists of 24M sentence pairs, including News Commentary, UN Parallel Corpus and CWMT Corpus.\footnote{http://www.statmt.org/wmt17/translation-task.html}
The newsdev2017 is used as the validation set and newstest2017 as the test set.
We also limit the maximal sentence length to 128.
For data pre-processing, we segment Chinese sentences with our in-house Chinese word segmentation tool and tokenize English sentences with the scripts provided in Moses.\footnote{https://github.com/moses-smt/mosesdecoder/blob/master\\/scripts/tokenizer/tokenizer.perl}
Then we learn a BPE model on pre-processed sentences with 32K merge operations, in which 44K and 33K sub-word tokens are adopted as source and target vocabularies respectively.


\paragraph{Experimental Details.} 
The Transformer model \citet{Vaswani2017AttentionIA} is adopted as our baseline.
For all translation tasks, we follow the \textit{transformer\_base\_v2} hyper-parameter setting\footnote{https://github.com/tensorflow/tensor2tensor/blob/v1.3.0/\\tensor2tensor/models/transformer.py} which corresponds to a 6-layer transformer with a model size of 512. 
The parameters are initialized using a normal distribution with a mean of 0 and a variance of $\sqrt{6/(d_{row}+d_{col})}$, where $d_{row}$ and $d_{col}$ are the number of rows and columns in the structure~\cite{glorot2010understanding}.
All models are trained on 4 Tesla M40 GPUs for a total of 100K steps using the Adam~\cite{Kingma2014AdamAM} algorithm.
The initial learning rate is set to 0.2 and decayed according to the schedule in \citet{Vaswani2017AttentionIA}.
During training, the batch size is set to approximately 4096 words per batch and checkpoints are created every 60 minutes.
At test time, we use a beam of 8 and a length penalty of 1.0.

Other hyper-parameters used in our approach are set as \( \lambda = 1, m = 1 \).
To build the synthetic data in Algorithm \ref{alg:regularization}, we adopt beam search to generate translation candidates with beam size 4, and the best sample is used for the estimation of KL divergence.
In practice, to speed up the decoding process, we sort all source sentences according to the sentence length, and then 32 sentences are simultaneously translated with parallel decoding implementation. 
In our experiments, we try different settings (\(\lambda=0.1,0.2,0.5,2.0,5.0\)), and find the $\lambda = 1$ achieves the best BLEU result on validation set.
We also test the model performance on validation set with the bigger parameter $m$. 
When $m=2,3,4$, we do not find the further improvement but it brings some training times due to more pseudo sentence pairs. 
In addition, we use sentence-level BLEU to filter wrong translations whose BLEU score is not greater than 30\%.
Notice that the R2L model gets comparable results with the L2R model, thus only the result of the L2R model is reported in our experiments.

\subsection{Evaluation on NIST Corpora}

Table \ref{table:chinese-english-NIST} shows the evaluation results of different models on NIST datasets. \textbf{MRT} represents \citet{shen-EtAl:2016:P16-1}'s method, \textbf{RT} denotes our regularization approach, and \textbf{JS} represents \citet{liu2016agreement}'s method that modifies the inference strategy by reranking the $N$-best results with a joint probability of bidirectional models. 
All the results are reported based on case-insensitive BLEU and computed using Moses \textit{multi-bleu.perl} script.

We observe that by taking agreement information into consideration, Transformer+JS and Transformer+RT can bring improvement across different test sets, in which our approach achieves 2.73 BLEU point improvements over Transformer on average.
These results confirm that introducing agreement between L2R and R2L models helps handle exposure bias problem and improves translation quality.

Besides, we see that Transformer+RT gains better performance than Transformer+JS across different test sets, with 1.5 BLEU point improvements on average.
Since Transformer+JS only leverages the agreement restriction in inference stage, L2R and R2L models still suffer from exposure bias problem and generate bad translation candidates, which limits the room for improvement in the re-ranking process.
Instead of combining R2L model during inference, our approach utilizes the intrinsic probabilistic connection between L2R and R2L models to guide the learning process. The two NMT models are expected to adjust in disagreement cases and then the exposure bias problem of them can be solved.

\begin{figure}[t] 
	\begin{center}
        \includegraphics[width = 3.3in]{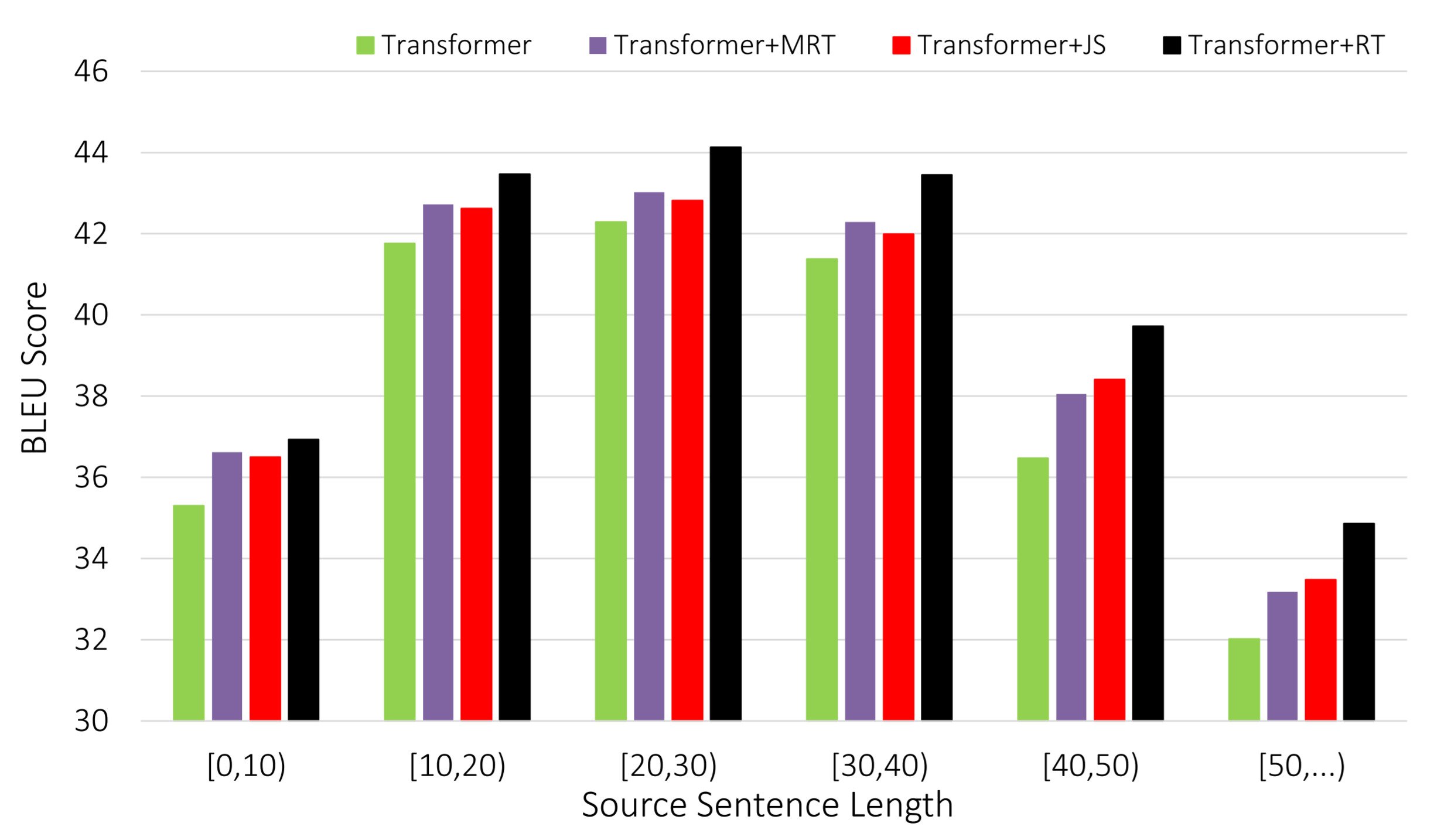}
	\end{center}
	\caption{Performance of the generated translations with respect to the length of source sentences on NIST datasets.}
	\label{fig:length-analysis}
\end{figure}

\begin{table*}[t]
	\centering
	\begin{tabular}{l|c|c|c|c}
		\hline
		& \multicolumn{2}{c|}{English-German} & \multicolumn{2}{c}{Chinese-English} \\ \cline{2-5} 
		System                            & newstest2016              & newstest2017             & newsdev2017               & newstest2017             \\ \hline \hline
		Transformer         & 32.58           & 25.48           & 20.87             & 23.01           \\ \hline
		Transformer+MRT     & 33.27                & 25.87                & 21.66                  & 24.24               \\ \hline
		Transformer+JS      &  32.91                & 25.93                & 21.25                  & 23.59                \\ \hline
		Transformer+RT                     & 34.56          & 27.18            & 22.50             & 25.38         \\ \hline \hline
		Transformer-big           &33.58         &27.13          &  21.91            & 24.03            \\ \hline
		Transformer-big+BT                & 35.06           & 28.34            &  23.59           & 25.53            \\ \hline
		Transformer-big+BT+RT & \textbf{36.78} & \textbf{29.46} & \textbf{24.84}  & \textbf{27.21} \\ \hline
		Edinburgh's NMT System (ensemble) & 36.20            & 28.30                & 24.00             & 25.70                \\ \hline
		Sogou's NMT System (ensemble)     & -                & -                & 22.90          & 26.40                \\ \hline
	\end{tabular}
	\caption{Case-sensitive BLEU scores (\%) for English-German and Chinese-English translation on WMT test sets. Edinburgh \cite{Sennrich2017TheUO} and Sogou \cite{Wang2017SogouNM} NMT systems are No.1 system in leaderboard of WMT 2017's English-German and Chinese-English translation tasks respectively.}
	\label{table:en-de-and-zh-en-WMT}
\end{table*}

Longer source sentence implies longer translation that more easily suffers from exposure bias problem.
To further verify our approach, we group source sentences of similar length together and calculate the BLEU score for each group. 
As shown in Figure \ref{fig:length-analysis}, we can see that our mechanism achieves the best performance in all groups. The gap between our method and the other three methods is small when the length is smaller than 10, and the gap becomes bigger when the sentences become longer. This further confirms the efficiency of our proposed method in dealing with the explore bias problem. 

\subsection{Evaluation on WMT17 Corpora}

For WMT17 Corpora, we verify the effectiveness of our approach on English-German and Chinese-English translation tasks from two angles: 
1) We compare our approach with baseline systems when only parallel corpora is used;
2) We investigate the impact of the combination of back-translation technique \cite{Sennrich2016ImprovingNM} and our approach.
Since back-translation method brings in more synthetic data, we choose Transformer-big settings defined in \citet{Vaswani2017AttentionIA} for this setting.
Experimental results are shown in Table \ref{table:en-de-and-zh-en-WMT}, in which \textbf{BT} denotes back-translation method.
In order to be comparable with NMT systems reported in WMT17, all results are reported based on case-sensitive BLEU and computed using official tools - SacreBLEU.\footnote{https://github.com/awslabs/sockeye/tree/master/contrib/\\sacrebleu}

\paragraph{Only Parallel Corpora.} 
As shown in Table \ref{table:en-de-and-zh-en-WMT}, Transformer+JS and Transformer+RT both generate improvements on test sets for English-German and Chinese-English translation tasks. 
It confirms the effectiveness of leveraging agreements between L2R and R2L models.
Additionally, our approach significantly outperforms Transformer+JS, yielding the best BLEU score on English-German and Chinese-English test sets respectively when merely using bilingual data.
These results further prove the effectiveness of our method.

\paragraph{Combining with Back-Translation Method.} 
To verify the effect of our approach when monolingual data is available, we combine our approach with back-translation method. 
We first randomly select 5M German sentences and 12M English sentences from ``News Crawl: articles from 2016". Then German-English and English-Chinese NMT systems learnt with parallel corpora are used to translate monolingual target sentences.

From Table \ref{table:en-de-and-zh-en-WMT}, we find that Transformer-big gains better performance than Transformer due to more model parameters.
When back-translation method is employed, Transformer-big+BT achieves 1.21 and 1.5 BLEU point improvements over Transformer-big on English-German and Chinese-English sets respectively.
Our method can further gain remarkable improvements based on back-translation method, resulting in the best results on all translation tasks.
These results show that in semi-supervised learning scenarios, the NMT model still can benefit from our proposed approach. 
In addition, our single model Transformer-big+BT+RT even achieves the best performance on WMT17's English-German and Chinese-English translation tasks, among all the reported results, including ensemble systems.

\subsection{Effect of Joint Training}

\begin{table}[t]
	\centering
	\begin{tabular}{c|c|c}
		\hline
		& English-German & Chinese-English \\ \hline \hline
		Iteration 0 & 32.58          & 20.87           \\ \hline
		Iteration 1 & 33.86          & 21.92           \\ \hline
		Iteration 2 & 34.56          & 22.50           \\ \hline
		Iteration 3 & 34.58          & 22.47           \\ \hline
	\end{tabular}
	\caption{Translation performance of our method on WMT validation sets during training process. ``Iteration 0'' denotes baseline Transformer model.}
	\label{table:joint-training}
\end{table}

We further investigate the impact of our joint training algorithm during the whole training process.
Table \ref{table:joint-training} shows the BLEU scores on WMT validation sets in each iteration.
For each iteration, we train NMT models until the performance on the development set does not increase.
We find that more iterations can lead to better evaluation results consistently and 2 iterations are enough to reach convergence in our experiments.
However, further iterations cannot bring noticeable translation accuracy improvements but more training time. 
For training cost, since our method is based on pre-trained models, the entire training time is almost 2 times the original MLE training.

\begin{table*}[t]
	\centering
	\begin{tabular}{c|l}
		\hline
		Source           & \makecell[l]{ \shou4\hai4\ren2 \de1 \ge1\ge1 Louis Galicia \dui4 \mei3\guo2 \guan3\bo1 \gong1\si1 \wei4\yu2 \jiu4\jin1\shan1 \de1 \dian4\tai2 KGO \biao3\shi4,\\ \zhi1\qian2 \zai4 \bo1\shi4\dun4 \zuo4 \liu2\shui3\xian4 \chu2\shi1 \de1 Frank \yu2 \liu4\ge4\yue4 \qian2 \zai4 \jiu4\jin1\shan1 \de1 Sons \&\\ Daughters \can1\guan1 \zhao3\dao4 \yi1\fen4 \liu2\shui3\xian4 \chu2\shi1 \de1 \li1\xiang3 \gong1\zuo4. } \\ \hline
		Reference        & \makecell[l]{  The victim's brother, Louis Galicia, told ABC station KGO in San Francisco that Frank, previously a\\ line cook in Boston, had landed his dream job as line chef at San Francisco’s Sons \& Daughters\\ restaurant six months ago.  }   \\ \hline \hline
		Transformer      & \makecell[l]{  Louis Galicia, the victim's brother, told ABC radio station KGO in San Francisco that Frank, who used\\ to work as an assembly line cook in Boston, \uwave{had found an ideal job in the Sons \& Daughters restaurant}\\ \uwave{in San Francisco} six months ago. }                               \\ \hline
		Transformer (R2L) & \makecell[l]{  The victim's brother, Louis Galia, told ABC's station KGO, \uwave{ in San Francisco, Frank} had found an ideal\\ job as a pipeline chef six months ago at Sons \& Daughters restaurant in San Francisco . }    \\ \hline
		Transformer+RT   & \makecell[l]{  The victim's brother, Louis Galicia, told ABC radio station KGO in San Francisco that Frank, who\\ previously worked as an assembly line cook in Boston, found an ideal job as an assembly line cook\\ six months ago at Sons \& Daughters restaurant in San Francisco. } \\ \hline
	\end{tabular}
	\caption{Translation examples of different systems. Text highlighted in wavy lines is incorrectly translated.}
	\label{table:example}
\end{table*}

\subsection{Example}

In this section, we give a case study to analyze our method. 
Table \ref{table:example} provides a Chinese-English translation example from newstest2017.
We find that Transformer produces the translation with good prefixes but bad suffixes, while Transformer (R2L) generates the translation with desirable suffixes but incorrect prefixes.
For our approach, we can see that Transformer+RT produces a high-quality translation in this case, which is much better than Transformer and Transformer (R2L). 
The reason is that leveraging the agreement between L2R and R2L models in training stage can better punish bad suffixes generated by Transformer and encourage desirable suffixes from Transformer (R2L).

\section{Related Work}

Target-bidirectional transduction techniques have been explored in statistical machine translation, under the IBM framework \cite{Watanabe2002BidirectionalDF} and the feature-driven linear models \cite{Finch2009BidirectionalPS,Zhang2013BeyondLM}.
Recently, \citet{liu2016agreement} and \citet{zhang2018AsynBid} migrate this method from SMT to NMT by modifying the inference strategy and decoder architecture of NMT.
\citet{liu2016agreement} propose to generate $N$-best translation candidates from L2R and R2L NMT models and leverage the joint probability of two models to find the best candidates from the combined $N$-best list.
\citet{zhang2018AsynBid} design a two-stage decoder architecture for NMT, which generates translation candidates in a right-to-left manner in first-stage and then gets final translation based on source sentence and previous generated R2L translation.
Different from their method, our approach directly exploits the target-bidirectional agreement in training stage by introducing regularization terms.
Without changing the neural network architecture and inference strategy, our method keeps the same speed as the original model during inference. 

To handle the exposure bias problem, many methods have been proposed, including designing new training objectives \cite{shen-EtAl:2016:P16-1,wiseman-rush:2016:EMNLP2016} and adopting reinforcement learning approaches \cite{ranzato2015sequence,bahdanau2016actor}.
\citet{shen-EtAl:2016:P16-1} attempt to directly minimize expected loss (maximize the expected BLEU) with Minimum Risk Training (MRT).
\citet{wiseman-rush:2016:EMNLP2016} adopt a beam-search optimization algorithm to reduce inconsistency between training and inference.
Besides, \citet{ranzato2015sequence} propose a mixture training method to perform a gradual transition from MLE training to BLEU score optimization using reinforcement learning.
\citet{bahdanau2016actor} design an actor-critic algorithm for sequence prediction, in which the NMT system is the actor, and a critic network is proposed to predict the value of output tokens.
Instead of designing task-specific objective functions or complex training strategies, our approach only adds regularization terms to standard training objective function, which is simple to implement but effective.


\section{Conclusion}

In this paper, we have presented a simple and efficient regularization approach to neural machine translation, which relies on the agreement between L2R and R2L NMT models. 
In our method, two Kullback-Leibler divergences based on probability distributions of L2R and R2L models are added to the standard training objective as regularization terms.
An efficient approximation algorithm is designed to enable fast training of the regularized training objective and then a training strategy is proposed to jointly optimize L2R and R2L models.
Empirical evaluations are conducted on Chinese-English and English-German translation tasks, demonstrating that our approach leads to significant improvements compared with strong baseline systems.

In our future work, we plan to test our method on other sequence-to-sequence tasks, such as summarization and dialogue generation.
Besides the back-translation method, it is also worth trying to integrate our approach with other semi-supervised methods to better leverage unlabeled data.

\section{Acknowledgments}

This research was partially supported by grants from the National Key Research and Development Program of China
(Grant No. 2018YFB1004300), the National Natural Science Foundation of China (Grant No. 61703386), the Anhui Provincial Natural Science Foundation (Grant No. 1708085QF140), and the Fundamental Research Funds for the Central Universities (Grant No. WK2150110006). 

Besides, we appreciate Dongdong Zhang and Ren Shuo for the fruitful discussions. We also thank the anonymous reviewers for their careful reading of our paper and insightful comments.

\bibliographystyle{aaai}
\bibliography{aaai19}

\end{document}